\pdfoutput=1

\documentclass[11pt]{article}

\usepackage[final]{emnlp2022}

\usepackage{times}
\usepackage{latexsym}

\usepackage[T1]{fontenc}

\usepackage[utf8]{inputenc}

\usepackage{microtype}
\usepackage{graphicx}
\usepackage{enumitem}
\usepackage{amsmath}

\newcommand{\ignore}[1]{\textcolor{gray}{}}
\newcommand*{\affmark}[1][*]{\textsuperscript{#1}}

\title{Automatic Document Selection for Efficient Encoder Pretraining}

\author{
Yukun Feng\affmark[1]~~~\, Patrick Xia\affmark[1]~~~\,  Benjamin Van Durme\affmark[1]~~~\,  João Sedoc\affmark[2]\\
{\affmark[1]Johns Hopkins University}\\
{\affmark[2]New York University }\\
\texttt{\{yfeng55, paxia, vandurme\}@jhu.edu, jsedoc@stern.nyu.edu}
}

\begin{document}
\maketitle

\begin{abstract}

Building pretrained language models is considered expensive and data-intensive, but must we increase dataset size to achieve better performance? 
We propose an alternative to larger training sets by automatically identifying smaller yet domain-representative subsets.
We extend \emph{Cynical Data Selection}, a statistical sentence scoring method that conditions on a representative target domain corpus. 
As an example, we treat the OntoNotes corpus as a target domain and pretrain a RoBERTa-like encoder from a cynically selected subset of the Pile.
On both 
perplexity and across several downstream tasks in the target domain, it consistently outperforms random selection with \textbf{20x} less data, \textbf{3x} fewer training iterations, and \textbf{2x} less estimated cloud compute cost, validating the recipe of automatic document selection for LM pretraining.
\end{abstract}

\section{Introduction}

Large pretrained language models have achieved state-of-the-art performance in NLP tasks~\cite[\textit{i.a.}]{devlin-etal-2019-bert,Liu2019RoBERTaAR}. 
These studies find that increasing pretraining data size usually leads to better task performance. 
For many tasks, additional task (in-domain) data helps improve the performance further~\citep{gururangan-etal-2020-dont, DBLP:journals/corr/abs-2109-07437, li-etal-2022-quantifying}. 
Several studies have found that directly pretraining on task data is more effective
: science texts~\citep{beltagy-etal-2019-scibert}, tweets~\citep{nguyen-etal-2020-bertweet}, legal texts~\cite{chalkidis-etal-2020-legal} or code~\citep{tabassum-etal-2020-code, chen2021codex}. 
Notably, these domains are known \textit{a priori}, and identifying data sources for curation is straightforward.
In other instances where the domain is less clear, like ``offensive online content'' \cite{bai-etal-2021-pre}, more complicated data sampling 
is employed to \textit{guess} at the desired data distribution suitable for training a downstream classifier.

\begin{figure}[t]
\includegraphics[scale=0.345]{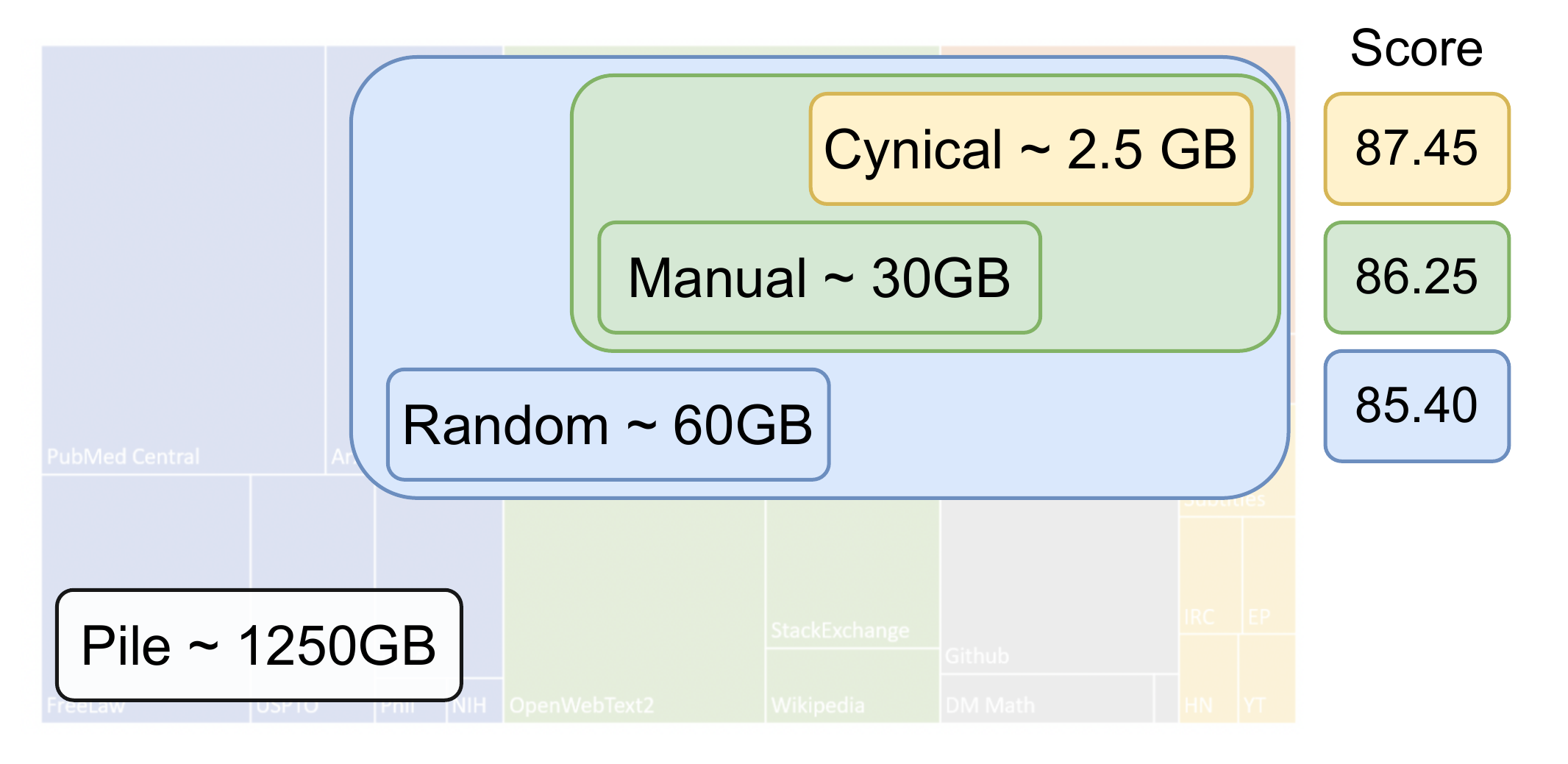}
\centering
\caption{This figure highlights the efficiency of the automatic cynical selection of documents in the target domain. Scores are averaged from 8 Edge Probing tasks. Cynically selected 2.5GB data achieves the best score.}
\label{fig-intro}
\end{figure}

To address such scenarios,  we propose automatically identifying relevant domain-specific training data for a large corpus and subsequently pretraining a model on the selected data. 
Specifically, we use Cynical Data Selection \citep{Axelrod2017}, an approach that advanced Moore-Lewis sampling~\citep{moore-lewis-2010-intelligent}, to select data from the Pile dataset~\citep{DBLP:journals/corr/abs-2101-00027}. 
This automatic selection method can include possibly overlooked yet relevant documents from domains that may not be too close to the target domain. 
\autoref{fig-intro} illustrates this method which achieves higher performance on tasks in the target domain by using only 2.5GB (0.5\%) of cynically selected data.

Specifically, we experiment with pretraining encoders with varying amounts of data sampled from the Pile.\footnote{The Pile consists of 800GB raw text but for this paper, we refer to its ``effective'' size, which is 1250GB.} 
With our ``target corpus'' of OntoNotes \cite{AB2/MKJJ2R_2013}, we compare language models trained with cynical and random selection at various data levels. 
We find that the cynically selected encoder achieves consistently lower target corpus perplexity than one trained with random selection. We further finetune the encoders on a suite of tasks, some of which are derived from OntoNotes. Again, we find that models pretrained with cynical selection perform best. We suggest this as a viable method for inexpensively pretraining effective domain-specific encoders.

\section{Cynical Data Selection}

Methods for data selection for language-related tasks have been widely studied, usually to select in-domain data \cite{axelrod-etal-2011-domain, van-der-wees-etal-2017-dynamic, https://doi.org/10.48550/arxiv.2010.01150, DBLP:journals/corr/abs-2012-10630}. One such method is Cynical Data Selection \cite{Axelrod2017}. The intuition behind cynical selection is greedily ranking sentences from the text corpus, based on its score computed against text \textit{representative} of the target domain, which is based on how much information gained by selecting it. 

Concretely, given representative text from the target domain, cynical selection uses the cross-entropy of the selected text against the representative text and calculates the information gain of each sentence in the general corpus. 
It then picks the most useful sentence relative to what has already been selected and its similarity to the representative text. 
This also leads to a bias towards shorter sentences and preferring sentences that contain words with high probability in the representative text.






Our work {\it extends} the cynical selection to the document level selection. Sentences are still scored at the sentence level, but the average sentence-level gain determines the information gain of a document.\footnote{A formal explanation of Cynical selection and its extension is in \autoref{appendix:cynds}.} We demonstrate its advantages in efficiently selecting related documents to the target domain.




\section{Experiments and Results}
In this work, we set OntoNotes 5.0 \cite{AB2/MKJJ2R_2013} as our target corpus, and we use a smaller sample from the training corpus of the CoNLL 2012 Shared Task \cite{pradhan-etal-2012-conll} as the representative corpus for data selection.
We first train an encoder based on the selected data and use the Edge Probing suite \cite{tenney2018what} for the downstream task evaluation, which has previously been used to probe and evaluate language models \cite{https://doi.org/10.48550/arxiv.1906.04341, https://doi.org/10.48550/arxiv.1905.05950, 
10.1162/tacl_a_00324, zhang-etal-2021-need}.

\begin{figure}[t]
\includegraphics[scale=0.48]{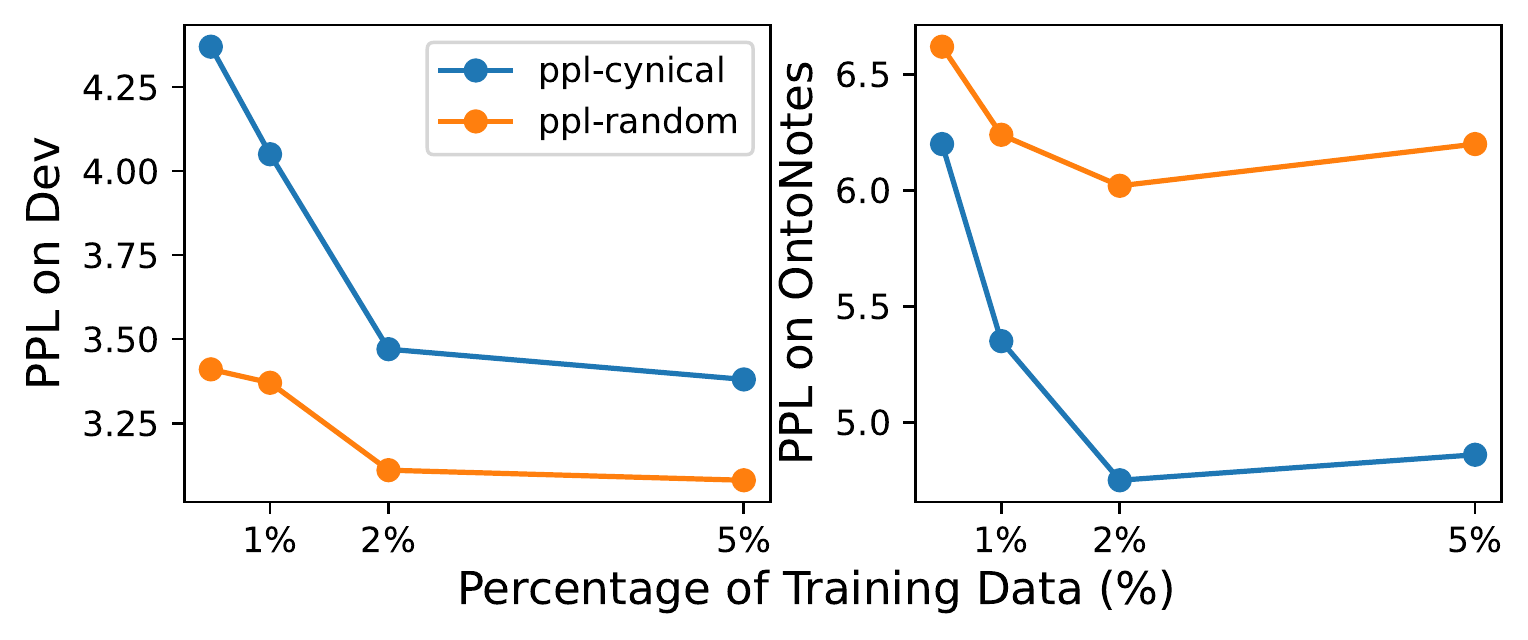}
\centering
\caption{Validation perplexity on held-out set (left), and OntoNotes (right) at 100k training steps.}
\label{fig-valid-ppl}
\end{figure}

\begin{figure*}[t]
\includegraphics[scale=0.58]{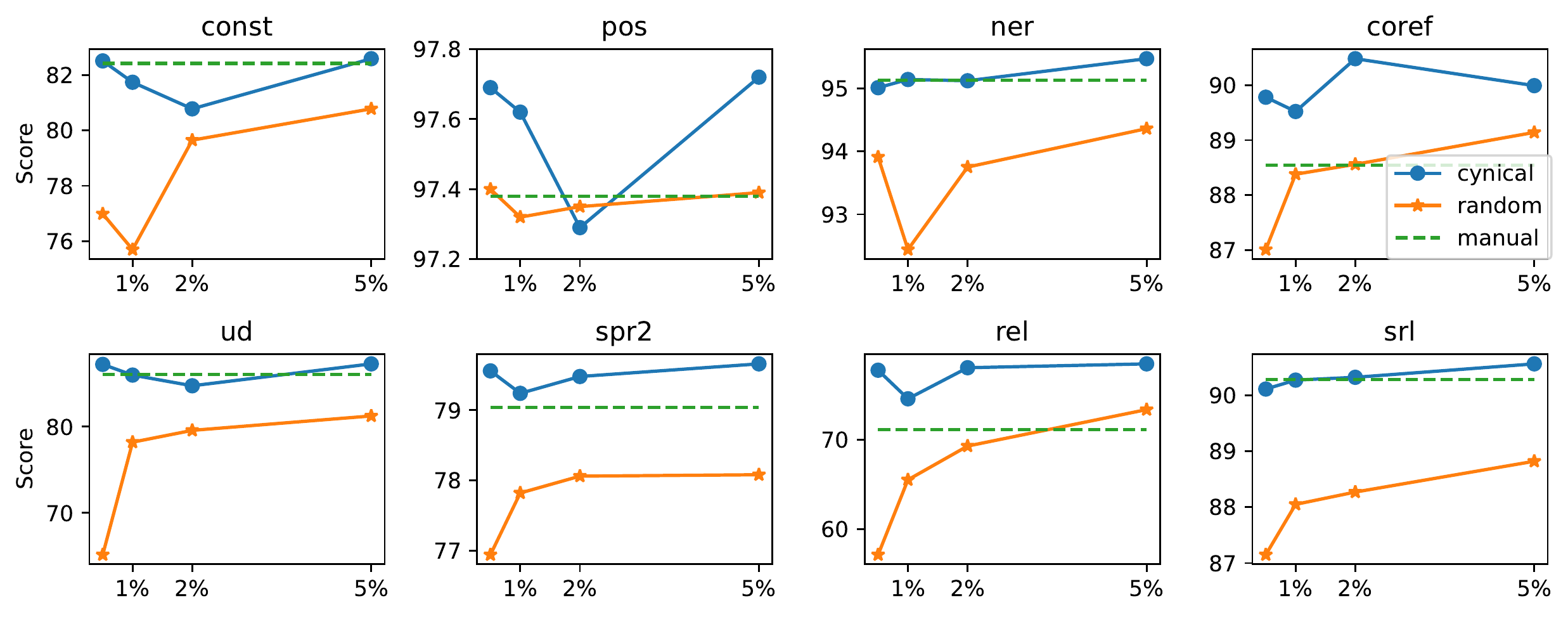}
\centering
\caption{Evaluation on 8 Edge Probing tasks \cite{tenney2018what}. The cynical selection consistently outperforms both the random and manual selection in most cases, even with only 0.5\% selected documents.
}
\label{fig-edge-eval}
\end{figure*}

\subsection{Data Selection}
\paragraph{Dataset}
We adopt the Pile \cite{DBLP:journals/corr/abs-2101-00027} for data selection, which consists of 1250GB text from 22 domains.
Cynical selection naturally prefers text data based on the target corpus.
To make a more fair comparison, we exclude 100GB data from ``DM Mathematics'' and ``Github'' to eliminate the noise of non-text data in random selection.

\paragraph{Selection Strategy}
Encoder pretraining is naturally a document-level task, as context contributes critically to improved representations. 
Thus, we need to extend the sentence selection into the document selection to achieve a better-contextualized representation at the pretraining stage.\footnote{We unsurprisingly find that selection at the document-level works better than at the sentence-level
(\autoref{sec:appendix}).}
We apply our extended document-level cynical selection to the Pile and extract the top $\{0.5\%, 1\%, 2\%, 5\%\}$ scored documents.\footnote{Our code repository is publicly available at \url{https://github.com/jsedoc/DL-CynDS}.}
We also randomly sample the same percentage of documents from Pile to use as a corresponding baseline. 
As a baseline for manual selection, we use 30GB text from "Wikipedia" and "BookCorpus" subsets, following \citet{Liu2019RoBERTaAR}.

\subsection{Encoder Pretraining}
We set up a BERT-base model and follow the pretraining objective and settings described in RoBERTa\cite{Liu2019RoBERTaAR}.\footnote{We adopt the training scripts from FairSeq for encoder pretraining, https://github.com/facebookresearch/fairseq.}
In \autoref{fig-valid-ppl}, we plot the validation perplexity on both the representative corpus (CoNLL 2012 Shared Task) and a held-out set of the Pile.
The perplexity on the held-out set decreases when there is more training data for both the cynical and random selection.
Cynical selection attains a higher perplexity, which shows that while the selected documents are more adapted to the target domain, it is not better adapted to the general corpus.
As each encoder needs different training steps for different corpus sizes, we try to make a fair comparison by assuming a fixed training budget of 
100k update steps. 
In Figure \ref{fig-valid-ppl}, we find that at 100k steps, 2\% of the cynically selected data achieves the lowest perplexity, and more training data does not help the adaptation to the target corpus.
Also,  cynical selected documents consistently outperforms the random selection, demonstrating the effectiveness of adapting to the target domain.

\subsection{Edge Probing Evaluation}
We evaluate the effectiveness of the pretrained encoders on 8 Edge Probing tasks \cite{tenney2018what},\footnote{We adopt the jiant for edge probing  data processing and finetuning, \url{https://github.com/nyu-mll/jiant}.}
for which the metric and architecture are uniformed to evaluate the span-level contextual representation of the language model, and it has been widely studied in the past few years.
Results are plotted in \autoref{fig-edge-eval}. We find:

\textbf{Observation 1:} Models trained on cynically selected documents show consistent performance gain on all tasks compared to the random selection.

\textbf{Observation 2:} In most tasks, even using only 0.5\% (2.5GB) of cynically selected documents
 outperforms the manually selected baseline (30GB).
 
\textbf{Observation 3:} Compared to random sampling, the performance gain of the cynical selected documents is larger with only 0.5\% to 1\% of training data, and  decreases for larger training sets as random selection catches up.

\textbf{Observation 4:} For some tasks, especially "const" and "pos," which are two tasks exactly based on the OntoNotes dataset, cynical selected documents yield good task performance with only 0.5\% data, and the scores decrease when increasing the selection size to 2\%, but increase again with 5\%.
This could suggest that in cynical selection, the top-scored documents are strongly related and helpful to the target task domain, while the others may not contribute as much or even hurt. However, more data ultimately does improve performance.

Overall, we could achieve promising results with only 0.5\% documents of the entire corpus, demonstrating the effectiveness and efficiency of cynical selection in the adaptation to downstream tasks in the target domain.
We also notice the standard deviation of the runs for random selection is 
much larger than cynical selection, indicating more stable encoder results from cynically selected documents.

\begin{figure}[t]
\includegraphics[scale=0.56]{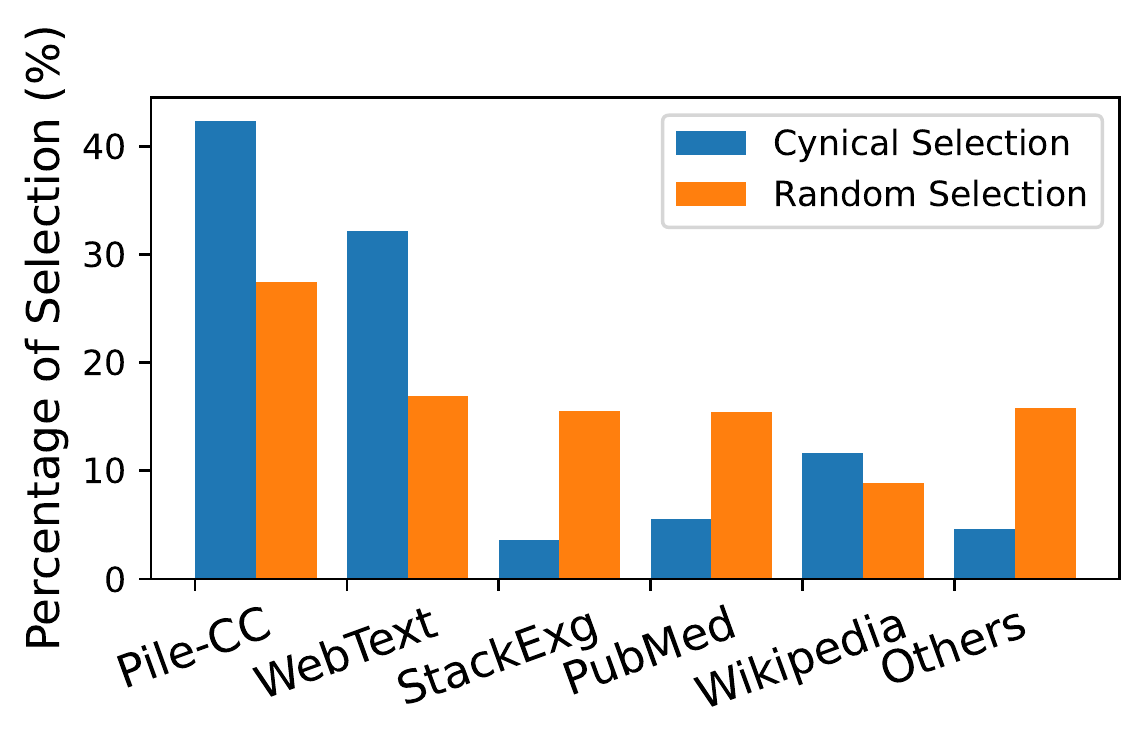}
\centering
\caption{Data distribution over the Pile domains}
\label{fig-distribution}
\end{figure}

\subsection{Discussion}
\begin{figure}[t]
\includegraphics[scale=0.56]{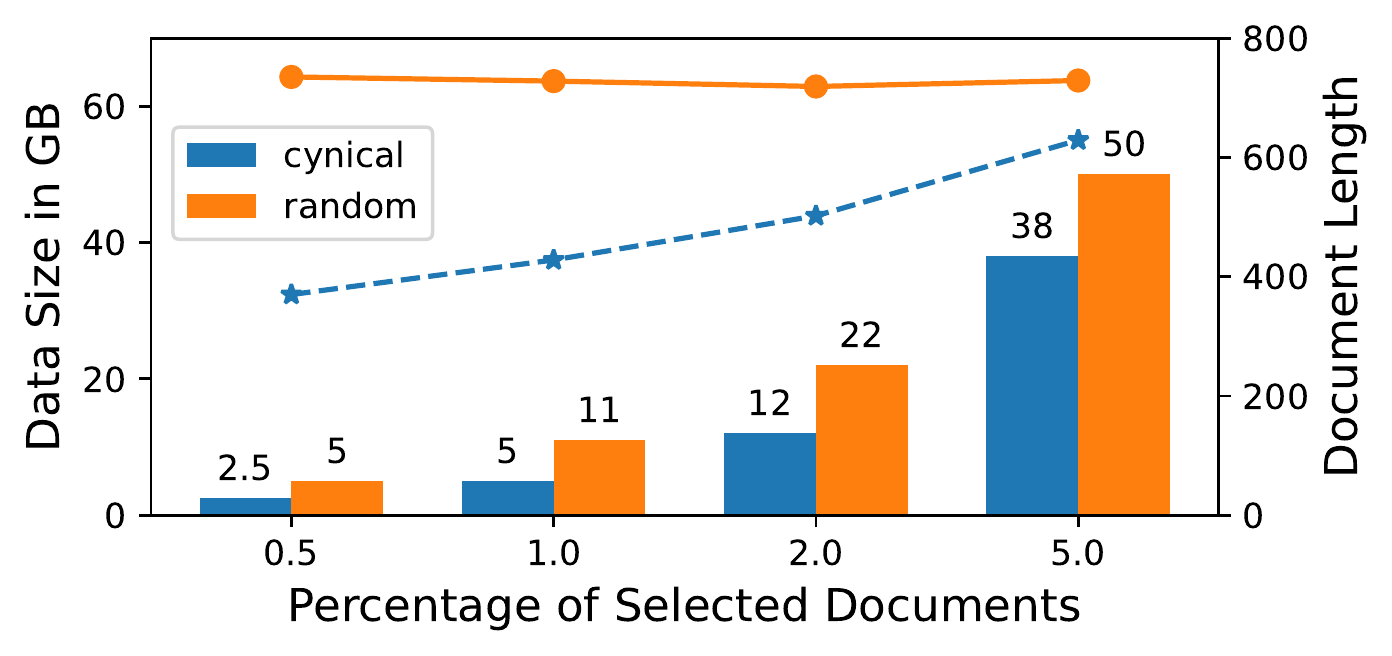}
\centering
\caption{For each percentage of cynically and randomly selected documents, we show the actual data size (GB) and corresponding document length. }
\label{fig-size}
\end{figure}

\paragraph{Data Distribution}
We plot the domain distribution of the selected documents in \autoref{fig-distribution}.
While random selection follows the distribution of the original Pile dataset,  cynical selection prefers news-like articles such as the "Pile CC" and "OpenWebText2," rather than technical ones, like StackExchange. 
Also, since we consider the same number of selected documents for each split, the actual selected data size is not the same (\autoref{fig-size}).
We notice that cynical selection prefers shorter documents, especially in the top-ranked samples.
This should be related to our scoring strategy since we average the sentence scores as the final document score.
In the case for long documents, even though there are sentences with higher scores, it is not very likely to be selected since the final scores are averaged by the total number of sentences.
This explains why the cynical selection prefers shorter documents in the 0.5\% and 1\% selection but not in the 5\% selection.
Therefore, when we bring the actual selected data sizes into the comparison, the cynical selection is much more efficient than the random sampling. Future work can investigate other methods of aggregating sentence-level scores.

\paragraph{Computational Trade-off}
Cynical selection enables the language models to use less training data and GPU time while achieving competitive results.
However, the data selection needs to be done before the training and pre-processing could be costly.
Cynical selection on 
the Pile can be parallelized via sharding, because the specific order/ranking of a document in the final selected subset is not important. The intuition is that any good document will be chosen early, regardless of which shard it is in. 
So, we split the automatic document selection of the Pile into 10,000 smaller jobs, each requiring a single core CPU\footnote{Intel Xeon E5-2620 v3, a chip from 2014.} and 10GB of RAM and taking 2 hours to finish.
In general, the cost of the selection depends on the size of the general corpus that is being selected from.
In our training environment with 8 RTX6000 GPUs, it takes 800+ GPU hours in total to train an encoder with 60GB randomly selected documents. 
To achieve comparable or even better performance with cynical selected documents, we only need 200 GPU hours for the 2.5GB of cynically selected data to converge.
The market price for a single RTX6000 is \$1.50/hour, so we need \$1200+ to train with random selection but less than \$300 for cynical selection.
On the Google Cloud Platform, 20,000 hours on comparable or faster CPUs can be obtained with \$200. 
Overall, cynical selected documents saves more than \textbf{50\%} of the computational cost and achieves better task scores.

\begin{figure}[t]
\includegraphics[scale=0.6]{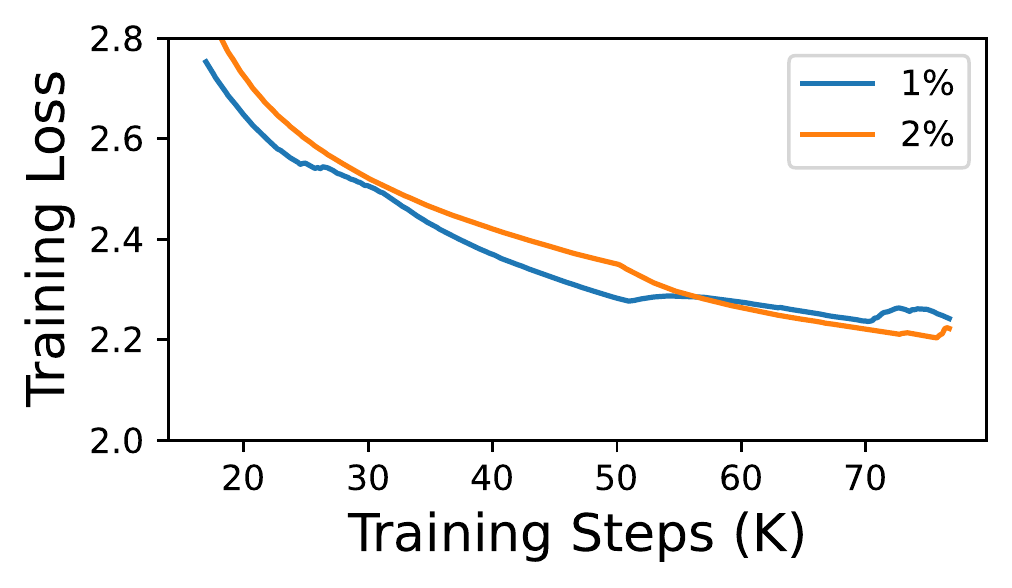}
\centering
\caption{This figure shows the training loss for the runs of 1\% and 2\% cynically selected subsets. 
}
\label{fig-loss}
\end{figure}

\paragraph{Overfitting}
Large language models have the ability to overfit or memorize small datasets \cite{kaplan2020scaling, Carlini2022QuantifyingMA}. We inspect the loss curves for two of the cynical selections (1\% and 2\%) in \autoref{fig-loss}. 
While the 1\% encoder achieves a lower loss for most parts of the training, it is eventually surpassed by the 2\% model.
This highlights a tradeoff between computing cost and performance; given a limited compute budget (in this example, under 50K steps), it is better to use a smaller selection. While prior work suggests scaling up models to fit dataset size \cite{kaplan2020scaling}, we are successful in \textit{scaling down} dataset sizes so that they can be efficiently fit (and outperform larger datasets) in fewer steps. 

\section{Related Work}
Due to the huge computational cost of training large models, both researchers and engineers have sought alternatives to using data more efficiently.
Some prior works use statistical methods to select relevant data from a large corpus \cite{Rousseau13,kirchhoff-bilmes-2014-submodularity, article, xu-koehn-2017-zipporah}.
Some other studies introduce additional classifiers or language models to help the data selection \cite{https://doi.org/10.48550/arxiv.1707.05246,10.1145/3289600.3290978, pmlr-v139-sun21a}.
Also, data selection could be more efficiently involved in the active learning approaches \cite{shen-etal-2004-multi, Lowell2018HowTA, erdmann-etal-2019-practical, 8983157, margatina-etal-2022-importance, https://doi.org/10.48550/arxiv.2205.03598}.
This work applies a \textbf{simple} statistical method to find the most related text to a target domain. It incrementally constructs a dataset out of a large corpus for the goal of training language models.

\section{Conclusion}
This work builds the connection from corpus subselection in statistical LM construction to neural LMs. 
We extend cynical data selection to efficiently select task-related documents for encoder pretraining and achieve lower perplexity in the target domain.
We also demonstrate its effectiveness on downstream tasks by achieving comparable or even better results with \textbf{20x} less data, \textbf{3x} fewer training iterations, and \textbf{2x} less computational cost on 8 Edge Probing tasks.
We believe this fills the gap in the literature on an important topic in training powerful LMs. 
We purposefully keep this work in the space of methods used in the days of Stat NLP to highlight their out-of-the-box applicability, for which that line of research is still salient.
Based on our findings, this line is resurrected, suggesting new novel approaches should be studied.
We anticipate that with this connection, researchers could explore this topic, investigate various subselection methods, and extend it to other domains.


\section*{Acknowledgements}
We thank all reviewers for their valuable feedback. 
We also appreciate the helpful suggestions from  Marc Marone, Amittai Axelrod, and Alex Warstadt.
This work is supported by IARPA BETTER (\#2019-19051600005).
The findings contained in this work are those of the authors and should not be interpreted as necessarily
representing the official policies, either expressed
or implied, or endorsements of IARPA or the U.S.
Government. The U.S. Government is authorized
to reproduce and distribute reprints for governmental purposes notwithstanding any copyright annotation therein.

\section*{Limitations}
Since pretraining encoders is expensive, our study only experiments on one source corpus (Pile) and one target task domain (OntoNotes). 
However, this method could be demonstrated more effectively on other datasets that are more domain-specific.
We do not run multiple random selections with different seeds due to the time and cost of training large models. We think the standard error for the randomly selected data would be significant, especially for the subset of only 0.5\% or 1\% documents. 
Also, we recognize that training our models longer or scaling up the model size is an ``easy'' method of improving performance \cite{Liu2019RoBERTaAR, kaplan2020scaling}. 
Our results assume a fixed training budget (max 100k steps). Thus with a larger budget, the trade-offs will vary.
Another concern is that we do not experiment with other subselection methods~\citep{gururangan-etal-2019-variational} or other languages, but we believe they should have similar trends.

\bibliography{anthology, custom}

\begin{thebibliography}{44}
\expandafter\ifx\csname natexlab\endcsname\relax\def\natexlab#1{#1}\fi

\bibitem[{Axelrod(2017)}]{Axelrod2017}
Amittai Axelrod. 2017.
\newblock \href
  {https://www.amazon.science/publications/cynical-selection-of-language-model-training-data}
  {Cynical selection of language model training data}.
\newblock \emph{arXiv}.

\bibitem[{Axelrod et~al.(2011)Axelrod, He, and Gao}]{axelrod-etal-2011-domain}
Amittai Axelrod, Xiaodong He, and Jianfeng Gao. 2011.
\newblock \href {https://aclanthology.org/D11-1033} {Domain adaptation via
  pseudo in-domain data selection}.
\newblock In \emph{Proceedings of the 2011 Conference on Empirical Methods in
  Natural Language Processing}, pages 355--362, Edinburgh, Scotland, UK.
  Association for Computational Linguistics.

\bibitem[{Bai et~al.(2021)Bai, Ritter, and Xu}]{bai-etal-2021-pre}
Fan Bai, Alan Ritter, and Wei Xu. 2021.
\newblock \href {https://doi.org/10.18653/v1/2021.emnlp-main.409} {Pre-train or
  annotate? domain adaptation with a constrained budget}.
\newblock In \emph{Proceedings of the 2021 Conference on Empirical Methods in
  Natural Language Processing}, pages 5002--5015, Online and Punta Cana,
  Dominican Republic. Association for Computational Linguistics.

\bibitem[{Beltagy et~al.(2019)Beltagy, Lo, and
  Cohan}]{beltagy-etal-2019-scibert}
Iz~Beltagy, Kyle Lo, and Arman Cohan. 2019.
\newblock \href {https://doi.org/10.18653/v1/D19-1371} {{S}ci{BERT}: A
  pretrained language model for scientific text}.
\newblock In \emph{Proceedings of the 2019 Conference on Empirical Methods in
  Natural Language Processing and the 9th International Joint Conference on
  Natural Language Processing (EMNLP-IJCNLP)}, pages 3615--3620, Hong Kong,
  China. Association for Computational Linguistics.

\bibitem[{Carlini et~al.(2022)Carlini, Ippolito, Jagielski, Lee, Tram{\`e}r,
  and Zhang}]{Carlini2022QuantifyingMA}
Nicholas Carlini, Daphne Ippolito, Matthew Jagielski, Katherine Lee, Florian
  Tram{\`e}r, and Chiyuan Zhang. 2022.
\newblock Quantifying memorization across neural language models.
\newblock \emph{ArXiv}, abs/2202.07646.

\bibitem[{Chalkidis et~al.(2020)Chalkidis, Fergadiotis, Malakasiotis, Aletras,
  and Androutsopoulos}]{chalkidis-etal-2020-legal}
Ilias Chalkidis, Manos Fergadiotis, Prodromos Malakasiotis, Nikolaos Aletras,
  and Ion Androutsopoulos. 2020.
\newblock \href {https://doi.org/10.18653/v1/2020.findings-emnlp.261}
  {{LEGAL}-{BERT}: The muppets straight out of law school}.
\newblock In \emph{Findings of the Association for Computational Linguistics:
  EMNLP 2020}, pages 2898--2904, Online. Association for Computational
  Linguistics.

\bibitem[{Chen et~al.(2021)Chen, Tworek, Jun, Yuan, de~Oliveira~Pinto, Kaplan,
  Edwards, Burda, Joseph, Brockman, Ray, Puri, Krueger, Petrov, Khlaaf, Sastry,
  Mishkin, Chan, Gray, Ryder, Pavlov, Power, Kaiser, Bavarian, Winter, Tillet,
  Such, Cummings, Plappert, Chantzis, Barnes, Herbert-Voss, Guss, Nichol,
  Paino, Tezak, Tang, Babuschkin, Balaji, Jain, Saunders, Hesse, Carr, Leike,
  Achiam, Misra, Morikawa, Radford, Knight, Brundage, Murati, Mayer, Welinder,
  McGrew, Amodei, McCandlish, Sutskever, and Zaremba}]{chen2021codex}
Mark Chen, Jerry Tworek, Heewoo Jun, Qiming Yuan, Henrique~Ponde
  de~Oliveira~Pinto, Jared Kaplan, Harri Edwards, Yuri Burda, Nicholas Joseph,
  Greg Brockman, Alex Ray, Raul Puri, Gretchen Krueger, Michael Petrov, Heidy
  Khlaaf, Girish Sastry, Pamela Mishkin, Brooke Chan, Scott Gray, Nick Ryder,
  Mikhail Pavlov, Alethea Power, Lukasz Kaiser, Mohammad Bavarian, Clemens
  Winter, Philippe Tillet, Felipe~Petroski Such, Dave Cummings, Matthias
  Plappert, Fotios Chantzis, Elizabeth Barnes, Ariel Herbert-Voss,
  William~Hebgen Guss, Alex Nichol, Alex Paino, Nikolas Tezak, Jie Tang, Igor
  Babuschkin, Suchir Balaji, Shantanu Jain, William Saunders, Christopher
  Hesse, Andrew~N. Carr, Jan Leike, Josh Achiam, Vedant Misra, Evan Morikawa,
  Alec Radford, Matthew Knight, Miles Brundage, Mira Murati, Katie Mayer, Peter
  Welinder, Bob McGrew, Dario Amodei, Sam McCandlish, Ilya Sutskever, and
  Wojciech Zaremba. 2021.
\newblock \href {http://arxiv.org/abs/2107.03374} {Evaluating large language
  models trained on code}.

\bibitem[{Clark et~al.(2019)Clark, Khandelwal, Levy, and
  Manning}]{https://doi.org/10.48550/arxiv.1906.04341}
Kevin Clark, Urvashi Khandelwal, Omer Levy, and Christopher~D. Manning. 2019.
\newblock \href {https://doi.org/10.48550/ARXIV.1906.04341} {What does bert
  look at? an analysis of bert's attention}.

\bibitem[{Dai et~al.(2020)Dai, Karimi, Hachey, and
  Paris}]{https://doi.org/10.48550/arxiv.2010.01150}
Xiang Dai, Sarvnaz Karimi, Ben Hachey, and Cecile Paris. 2020.
\newblock \href {https://doi.org/10.48550/ARXIV.2010.01150} {Cost-effective
  selection of pretraining data: A case study of pretraining bert on social
  media}.

\bibitem[{Dery et~al.(2021)Dery, Michel, Talwalkar, and
  Neubig}]{DBLP:journals/corr/abs-2109-07437}
Lucio~M. Dery, Paul Michel, Ameet Talwalkar, and Graham Neubig. 2021.
\newblock \href {http://arxiv.org/abs/2109.07437} {Should we be pre-training?
  an argument for end-task aware training as an alternative}.
\newblock \emph{CoRR}, abs/2109.07437.

\bibitem[{Devlin et~al.(2019)Devlin, Chang, Lee, and
  Toutanova}]{devlin-etal-2019-bert}
Jacob Devlin, Ming-Wei Chang, Kenton Lee, and Kristina Toutanova. 2019.
\newblock \href {https://doi.org/10.18653/v1/N19-1423} {{BERT}: Pre-training of
  deep bidirectional transformers for language understanding}.
\newblock In \emph{Proceedings of the 2019 Conference of the North {A}merican
  Chapter of the Association for Computational Linguistics: Human Language
  Technologies, Volume 1 (Long and Short Papers)}, pages 4171--4186,
  Minneapolis, Minnesota. Association for Computational Linguistics.

\bibitem[{Eetemadi et~al.(2015)Eetemadi, Lewis, Toutanova, and Radha}]{article}
Sauleh Eetemadi, William Lewis, Kristina Toutanova, and Hayder Radha. 2015.
\newblock \href {https://doi.org/10.1007/s10590-015-9176-1} {Survey of
  data-selection methods in statistical machine translation}.
\newblock \emph{Machine Translation}, 29.

\bibitem[{Erdmann et~al.(2019)Erdmann, Wrisley, Allen, Brown,
  Cohen-Bod{\'e}n{\`e}s, Elsner, Feng, Joseph, Joyeux-Prunel, and
  de~Marneffe}]{erdmann-etal-2019-practical}
Alexander Erdmann, David~Joseph Wrisley, Benjamin Allen, Christopher Brown,
  Sophie Cohen-Bod{\'e}n{\`e}s, Micha Elsner, Yukun Feng, Brian Joseph,
  B{\'e}atrice Joyeux-Prunel, and Marie-Catherine de~Marneffe. 2019.
\newblock \href {https://doi.org/10.18653/v1/N19-1231} {Practical, efficient,
  and customizable active learning for named entity recognition in the digital
  humanities}.
\newblock In \emph{Proceedings of the 2019 Conference of the North {A}merican
  Chapter of the Association for Computational Linguistics: Human Language
  Technologies, Volume 1 (Long and Short Papers)}, pages 2223--2234,
  Minneapolis, Minnesota. Association for Computational Linguistics.

\bibitem[{Gao et~al.(2021)Gao, Biderman, Black, Golding, Hoppe, Foster, Phang,
  He, Thite, Nabeshima, Presser, and Leahy}]{DBLP:journals/corr/abs-2101-00027}
Leo Gao, Stella Biderman, Sid Black, Laurence Golding, Travis Hoppe, Charles
  Foster, Jason Phang, Horace He, Anish Thite, Noa Nabeshima, Shawn Presser,
  and Connor Leahy. 2021.
\newblock \href {https://arxiv.org/abs/2101.00027} {The pile: An 800gb dataset
  of diverse text for language modeling}.
\newblock \emph{CoRR}, abs/2101.00027.

\bibitem[{Gururangan et~al.(2019)Gururangan, Dang, Card, and
  Smith}]{gururangan-etal-2019-variational}
Suchin Gururangan, Tam Dang, Dallas Card, and Noah~A. Smith. 2019.
\newblock \href {https://doi.org/10.18653/v1/P19-1590} {Variational pretraining
  for semi-supervised text classification}.
\newblock In \emph{Proceedings of the 57th Annual Meeting of the Association
  for Computational Linguistics}, pages 5880--5894, Florence, Italy.
  Association for Computational Linguistics.

\bibitem[{Gururangan et~al.(2020)Gururangan, Marasovi{\'c}, Swayamdipta, Lo,
  Beltagy, Downey, and Smith}]{gururangan-etal-2020-dont}
Suchin Gururangan, Ana Marasovi{\'c}, Swabha Swayamdipta, Kyle Lo, Iz~Beltagy,
  Doug Downey, and Noah~A. Smith. 2020.
\newblock \href {https://doi.org/10.18653/v1/2020.acl-main.740} {Don{'}t stop
  pretraining: Adapt language models to domains and tasks}.
\newblock In \emph{Proceedings of the 58th Annual Meeting of the Association
  for Computational Linguistics}, pages 8342--8360, Online. Association for
  Computational Linguistics.

\bibitem[{Hendrickx et~al.(2010)Hendrickx, Kim, Kozareva, Nakov,
  {\'O}~S{\'e}aghdha, Pad{\'o}, Pennacchiotti, Romano, and
  Szpakowicz}]{hendrickx-etal-2010-semeval}
Iris Hendrickx, Su~Nam Kim, Zornitsa Kozareva, Preslav Nakov, Diarmuid
  {\'O}~S{\'e}aghdha, Sebastian Pad{\'o}, Marco Pennacchiotti, Lorenza Romano,
  and Stan Szpakowicz. 2010.
\newblock \href {https://aclanthology.org/S10-1006} {{S}em{E}val-2010 task 8:
  Multi-way classification of semantic relations between pairs of nominals}.
\newblock In \emph{Proceedings of the 5th International Workshop on Semantic
  Evaluation}, pages 33--38, Uppsala, Sweden. Association for Computational
  Linguistics.

\bibitem[{Jiang et~al.(2020)Jiang, Xu, Araki, and
  Neubig}]{10.1162/tacl_a_00324}
Zhengbao Jiang, Frank~F. Xu, Jun Araki, and Graham Neubig. 2020.
\newblock \href {https://doi.org/10.1162/tacl_a_00324} {{How Can We Know What
  Language Models Know?}}
\newblock \emph{Transactions of the Association for Computational Linguistics},
  8:423--438.

\bibitem[{Kaplan et~al.(2020)Kaplan, McCandlish, Henighan, Brown, Chess, Child,
  Gray, Radford, Wu, and Amodei}]{kaplan2020scaling}
Jared Kaplan, Sam McCandlish, Tom Henighan, Tom~B. Brown, Benjamin Chess, Rewon
  Child, Scott Gray, Alec Radford, Jeffrey Wu, and Dario Amodei. 2020.
\newblock \href {http://arxiv.org/abs/2001.08361} {Scaling laws for neural
  language models}.

\bibitem[{Killamsetty et~al.(2020)Killamsetty, Sivasubramanian, Ramakrishnan,
  and Iyer}]{DBLP:journals/corr/abs-2012-10630}
KrishnaTeja Killamsetty, Durga Sivasubramanian, Ganesh Ramakrishnan, and
  Rishabh~K. Iyer. 2020.
\newblock \href {http://arxiv.org/abs/2012.10630} {{GLISTER:} generalization
  based data subset selection for efficient and robust learning}.
\newblock \emph{CoRR}, abs/2012.10630.

\bibitem[{Kirchhoff and Bilmes(2014)}]{kirchhoff-bilmes-2014-submodularity}
Katrin Kirchhoff and Jeff Bilmes. 2014.
\newblock \href {https://doi.org/10.3115/v1/D14-1014} {Submodularity for data
  selection in machine translation}.
\newblock In \emph{Proceedings of the 2014 Conference on Empirical Methods in
  Natural Language Processing ({EMNLP})}, pages 131--141, Doha, Qatar.
  Association for Computational Linguistics.

\bibitem[{Li et~al.(2022)Li, Yu, Khabsa, Zettlemoyer, Halevy, and
  Andreas}]{li-etal-2022-quantifying}
Belinda Li, Jane Yu, Madian Khabsa, Luke Zettlemoyer, Alon Halevy, and Jacob
  Andreas. 2022.
\newblock \href {https://doi.org/10.18653/v1/2022.naacl-main.346} {Quantifying
  adaptability in pre-trained language models with 500 tasks}.
\newblock In \emph{Proceedings of the 2022 Conference of the North American
  Chapter of the Association for Computational Linguistics: Human Language
  Technologies}, pages 4696--4715, Seattle, United States. Association for
  Computational Linguistics.

\bibitem[{Liu et~al.(2019)Liu, Ott, Goyal, Du, Joshi, Chen, Levy, Lewis,
  Zettlemoyer, and Stoyanov}]{Liu2019RoBERTaAR}
Yinhan Liu, Myle Ott, Naman Goyal, Jingfei Du, Mandar Joshi, Danqi Chen, Omer
  Levy, Mike Lewis, Luke Zettlemoyer, and Veselin Stoyanov. 2019.
\newblock Roberta: A robustly optimized bert pretraining approach.
\newblock \emph{ArXiv}, abs/1907.11692.

\bibitem[{Lowell et~al.(2018)Lowell, Lipton, and Wallace}]{Lowell2018HowTA}
David Lowell, Zachary~Chase Lipton, and Byron~C. Wallace. 2018.
\newblock How transferable are the datasets collected by active learners?
\newblock \emph{ArXiv}, abs/1807.04801.

\bibitem[{Margatina et~al.(2022)Margatina, Barrault, and
  Aletras}]{margatina-etal-2022-importance}
Katerina Margatina, Loic Barrault, and Nikolaos Aletras. 2022.
\newblock \href {https://aclanthology.org/2022.acl-short.93} {On the importance
  of effectively adapting pretrained language models for active learning}.
\newblock In \emph{Proceedings of the 60th Annual Meeting of the Association
  for Computational Linguistics (Volume 2: Short Papers)}, pages 825--836,
  Dublin, Ireland. Association for Computational Linguistics.

\bibitem[{Moore and Lewis(2010)}]{moore-lewis-2010-intelligent}
Robert~C. Moore and William Lewis. 2010.
\newblock \href {https://aclanthology.org/P10-2041} {Intelligent selection of
  language model training data}.
\newblock In \emph{Proceedings of the {ACL} 2010 Conference Short Papers},
  pages 220--224, Uppsala, Sweden. Association for Computational Linguistics.

\bibitem[{Nguyen et~al.(2020)Nguyen, Vu, and
  Tuan~Nguyen}]{nguyen-etal-2020-bertweet}
Dat~Quoc Nguyen, Thanh Vu, and Anh Tuan~Nguyen. 2020.
\newblock \href {https://doi.org/10.18653/v1/2020.emnlp-demos.2} {{BERT}weet: A
  pre-trained language model for {E}nglish tweets}.
\newblock In \emph{Proceedings of the 2020 Conference on Empirical Methods in
  Natural Language Processing: System Demonstrations}, pages 9--14, Online.
  Association for Computational Linguistics.

\bibitem[{Pradhan et~al.(2012)Pradhan, Moschitti, Xue, Uryupina, and
  Zhang}]{pradhan-etal-2012-conll}
Sameer Pradhan, Alessandro Moschitti, Nianwen Xue, Olga Uryupina, and Yuchen
  Zhang. 2012.
\newblock \href {https://aclanthology.org/W12-4501} {{C}o{NLL}-2012 shared
  task: Modeling multilingual unrestricted coreference in {O}nto{N}otes}.
\newblock In \emph{Joint Conference on {EMNLP} and {C}o{NLL} - Shared Task},
  pages 1--40, Jeju Island, Korea. Association for Computational Linguistics.

\bibitem[{Qu et~al.(2019)Qu, Ji, Qiu, Yang, Min, Chen, Huang, and
  Croft}]{10.1145/3289600.3290978}
Chen Qu, Feng Ji, Minghui Qiu, Liu Yang, Zhiyu Min, Haiqing Chen, Jun Huang,
  and W.~Bruce Croft. 2019.
\newblock \href {https://doi.org/10.1145/3289600.3290978} {Learning to
  selectively transfer: Reinforced transfer learning for deep text matching}.
\newblock In \emph{Proceedings of the Twelfth ACM International Conference on
  Web Search and Data Mining}, WSDM '19, page 699–707, New York, NY, USA.
  Association for Computing Machinery.

\bibitem[{Rousseau(2013)}]{Rousseau13}
Anthony Rousseau. 2013.
\newblock Xenc: An open-source tool for data selection in natural language
  processing.
\newblock \emph{The Prague Bulletin of Mathematical Linguistics}, (100):73--82.

\bibitem[{Ruder and Plank(2017)}]{https://doi.org/10.48550/arxiv.1707.05246}
Sebastian Ruder and Barbara Plank. 2017.
\newblock \href {https://doi.org/10.48550/ARXIV.1707.05246} {Learning to select
  data for transfer learning with bayesian optimization}.

\bibitem[{Rudinger et~al.(2018)Rudinger, Teichert, Culkin, Zhang, and
  Van~Durme}]{rudinger-etal-2018-neural}
Rachel Rudinger, Adam Teichert, Ryan Culkin, Sheng Zhang, and Benjamin
  Van~Durme. 2018.
\newblock \href {https://doi.org/10.18653/v1/D18-1114} {Neural-{D}avidsonian
  semantic proto-role labeling}.
\newblock In \emph{Proceedings of the 2018 Conference on Empirical Methods in
  Natural Language Processing}, pages 944--955, Brussels, Belgium. Association
  for Computational Linguistics.

\bibitem[{Shelmanov et~al.(2019)Shelmanov, Liventsev, Kireev, Khromov,
  Panchenko, Fedulova, and Dylov}]{8983157}
Artem Shelmanov, Vadim Liventsev, Danil Kireev, Nikita Khromov, Alexander
  Panchenko, Irina Fedulova, and Dmitry~V. Dylov. 2019.
\newblock \href {https://doi.org/10.1109/BIBM47256.2019.8983157} {Active
  learning with deep pre-trained models for sequence tagging of clinical and
  biomedical texts}.
\newblock In \emph{2019 IEEE International Conference on Bioinformatics and
  Biomedicine (BIBM)}, pages 482--489.

\bibitem[{Shen et~al.(2004)Shen, Zhang, Su, Zhou, and
  Tan}]{shen-etal-2004-multi}
Dan Shen, Jie Zhang, Jian Su, Guodong Zhou, and Chew-Lim Tan. 2004.
\newblock \href {https://doi.org/10.3115/1218955.1219030} {Multi-criteria-based
  active learning for named entity recognition}.
\newblock In \emph{Proceedings of the 42nd Annual Meeting of the Association
  for Computational Linguistics ({ACL}-04)}, pages 589--596, Barcelona, Spain.

\bibitem[{Silveira et~al.(2014)Silveira, Dozat, de~Marneffe, Bowman, Connor,
  Bauer, and Manning}]{silveira-etal-2014-gold}
Natalia Silveira, Timothy Dozat, Marie-Catherine de~Marneffe, Samuel Bowman,
  Miriam Connor, John Bauer, and Chris Manning. 2014.
\newblock \href
  {http://www.lrec-conf.org/proceedings/lrec2014/pdf/1089_Paper.pdf} {A gold
  standard dependency corpus for {E}nglish}.
\newblock In \emph{Proceedings of the Ninth International Conference on
  Language Resources and Evaluation ({LREC}'14)}, pages 2897--2904, Reykjavik,
  Iceland. European Language Resources Association (ELRA).

\bibitem[{Sun et~al.(2021)Sun, Dou, Li, Yan, Ouyang, and
  Cui}]{pmlr-v139-sun21a}
Ming Sun, Haoxuan Dou, Baopu Li, Junjie Yan, Wanli Ouyang, and Lei Cui. 2021.
\newblock \href {https://proceedings.mlr.press/v139/sun21a.html} {Autosampling:
  Search for effective data sampling schedules}.
\newblock In \emph{Proceedings of the 38th International Conference on Machine
  Learning}, volume 139 of \emph{Proceedings of Machine Learning Research},
  pages 9923--9933. PMLR.

\bibitem[{Tabassum et~al.(2020)Tabassum, Maddela, Xu, and
  Ritter}]{tabassum-etal-2020-code}
Jeniya Tabassum, Mounica Maddela, Wei Xu, and Alan Ritter. 2020.
\newblock \href {https://doi.org/10.18653/v1/2020.acl-main.443} {Code and named
  entity recognition in {S}tack{O}verflow}.
\newblock In \emph{Proceedings of the 58th Annual Meeting of the Association
  for Computational Linguistics}, pages 4913--4926, Online. Association for
  Computational Linguistics.

\bibitem[{Tenney et~al.(2019{\natexlab{a}})Tenney, Das, and
  Pavlick}]{https://doi.org/10.48550/arxiv.1905.05950}
Ian Tenney, Dipanjan Das, and Ellie Pavlick. 2019{\natexlab{a}}.
\newblock \href {https://doi.org/10.48550/ARXIV.1905.05950} {Bert rediscovers
  the classical nlp pipeline}.

\bibitem[{Tenney et~al.(2019{\natexlab{b}})Tenney, Xia, Chen, Wang, Poliak,
  McCoy, Kim, Durme, Bowman, Das, and Pavlick}]{tenney2018what}
Ian Tenney, Patrick Xia, Berlin Chen, Alex Wang, Adam Poliak, R~Thomas McCoy,
  Najoung Kim, Benjamin~Van Durme, Sam Bowman, Dipanjan Das, and Ellie Pavlick.
  2019{\natexlab{b}}.
\newblock \href {https://openreview.net/forum?id=SJzSgnRcKX} {What do you learn
  from context? probing for sentence structure in contextualized word
  representations}.
\newblock In \emph{International Conference on Learning Representations}.

\bibitem[{Tsvigun et~al.(2022)Tsvigun, Shelmanov, Kuzmin, Sanochkin, Larionov,
  Gusev, Avetisian, and Zhukov}]{https://doi.org/10.48550/arxiv.2205.03598}
Akim Tsvigun, Artem Shelmanov, Gleb Kuzmin, Leonid Sanochkin, Daniil Larionov,
  Gleb Gusev, Manvel Avetisian, and Leonid Zhukov. 2022.
\newblock \href {https://doi.org/10.48550/ARXIV.2205.03598} {Towards
  computationally feasible deep active learning}.

\bibitem[{van~der Wees et~al.(2017)van~der Wees, Bisazza, and
  Monz}]{van-der-wees-etal-2017-dynamic}
Marlies van~der Wees, Arianna Bisazza, and Christof Monz. 2017.
\newblock \href {https://doi.org/10.18653/v1/D17-1147} {Dynamic data selection
  for neural machine translation}.
\newblock In \emph{Proceedings of the 2017 Conference on Empirical Methods in
  Natural Language Processing}, pages 1400--1410, Copenhagen, Denmark.
  Association for Computational Linguistics.

\bibitem[{Weischedel et~al.(2013)Weischedel, Palmer, Marcus, Hovy, Pradhan,
  Ramshaw, Xue, Taylor, Kaufman, Franchini, El-Bachouti, Belvin, and
  Houston}]{AB2/MKJJ2R_2013}
Ralph Weischedel, Martha Palmer, Mitchell Marcus, Eduard Hovy, Sameer Pradhan,
  Lance Ramshaw, Nianwen Xue, Ann Taylor, Jeff Kaufman, Michelle Franchini,
  Mohammed El-Bachouti, Robert Belvin, and Ann Houston. 2013.
\newblock \href {https://doi.org/11272.1/AB2/MKJJ2R} {{OntoNotes Release 5.0}}.

\bibitem[{Xu and Koehn(2017)}]{xu-koehn-2017-zipporah}
Hainan Xu and Philipp Koehn. 2017.
\newblock \href {https://doi.org/10.18653/v1/D17-1319} {{Z}ipporah: a fast and
  scalable data cleaning system for noisy web-crawled parallel corpora}.
\newblock In \emph{Proceedings of the 2017 Conference on Empirical Methods in
  Natural Language Processing}, pages 2945--2950, Copenhagen, Denmark.
  Association for Computational Linguistics.

\bibitem[{Zhang et~al.(2021)Zhang, Warstadt, Li, and
  Bowman}]{zhang-etal-2021-need}
Yian Zhang, Alex Warstadt, Xiaocheng Li, and Samuel~R. Bowman. 2021.
\newblock \href {https://doi.org/10.18653/v1/2021.acl-long.90} {When do you
  need billions of words of pretraining data?}
\newblock In \emph{Proceedings of the 59th Annual Meeting of the Association
  for Computational Linguistics and the 11th International Joint Conference on
  Natural Language Processing (Volume 1: Long Papers)}, pages 1112--1125,
  Online. Association for Computational Linguistics.

\end{thebibliography}
\bibliographystyle{acl_natbib}

\clearpage
\appendix

\section{Appendix}
\label{sec:appendix}

\subsection{Detailed Distribution}
A detailed data distribution is shown in \autoref{detailed-data-distribution}.

\section{Formalization of Cynical Data Selection}
\label{appendix:cynds}

The aim of CynDS is to incrementally construct $W$ through scoring each sentence by information gained relative to the already selected data  (\autoref{eqn:info-gain}).

Given a \textit{REP}resentative corpus from the target domain, CynDS is an effective and efficient method to identify the most relevant subset of sentences from a large corpus.
Formally, we can define a cross-entropy between REP and some set of tokens as,
\begin{equation*}
H(REP) = - \sum_{v\in{V_{REP}}} \frac{C_{REP}(v)}{W_{REP}}log\frac{C(v)}{|W|},
\end{equation*}
where $W$ is the set of tokens, $V$ is the vocabulary, and $C$ indicates the count of word type, $v$. $C_\text{REP}(v)$ is the count within REP and $C(v)$ is the count within $W$.

Let $W_1, \ldots, W_n$ be the incrementally selected corpus. We can define the cross-entropy after selecting $n$ sentences as
\begin{align*}
& H_n(REP) = - \sum_{v\in{V_{REP}}} \frac{C_{REP}(v)}{W_{REP}}log\frac{C_n(v)}{W_n}
\end{align*}
and minimize $H_n$. This can be rewritten recursively as 
\begin{align*}
\label{increment}
& H_{n+1} = H_{n} + \max_{s}  \underset{n\to{n+1}}{\Delta H}(s)
\end{align*}
where $\underset{n\to{n+1}}{\Delta H}(s)$ is the delta (effect) of a given sentence $s$.
To find the ${n+1}^{th}$ sentence that minimizes $\underset{n\to{n+1}}{\Delta H}$, we can rewrite it as
\begin{equation}
\underset{n\to{n+1}}{\Delta H} = \underset{n\to{n+1}}{Penalty} + \underset{n\to{n+1}}{Gain}
\label{eqn:info-gain}
\end{equation}

Here, penalty refers to how similar the sentence is to already selected texts and gain refers to how similar the sentence is to the representative corpus. \citet{Axelrod2017} derives the $Penalty$ and $Gain$ as 
\begin{align*}
&\underset{n\to{n+1}}{Penalty} = log\frac{|W_n + w_{n+1}|}{|W_n|} \\
&\underset{n\to{n+1}}{Gain} = \sum_{v\in V_{REP}}\frac{C_{REP}(v)}{W_{REP}}log\frac{C_n(v)}{C_n(v)+c_{n+1}(v)}
\end{align*}

A proof of this derivation is given in \citet{Axelrod2017}.

In our work, we still let $W_1, \ldots, W_n$ represent the first $n$ sentences, and $H(REP)$ is unchanged. However, we use the scores, $\underset{n\to{n+1}}{\Delta H}(s)$, of each sentence and compute document-level scores for each document, 
\begin{align*}
    Score(D) = \frac{1}{|D|}\sum_{s\in D}\underset{n\to{n+1}}{\Delta H}(s)
\end{align*}

These document-level scores can then be ranked, and we select the top $k\%$ of the documents. Note that while there are many alternatives to selecting documents, our goal is to select \textit{a} method and evaluate whether automatic data selection is effective for LM pretraining rather than comparing different methods, which can be future work.



\subsection{Sentence vs Document Selection}
Results are shown below in \autoref{sent-doc}.

\begin{table}[!ht]
\centering
\begin{tabular}{cc}
\hline
Data & ppl on OntoNotes\\ 
\hline
Cynical Sent & 102.21 \\
Cynical Doc & 4.98 \\
Random Doc & 8.77 \\
\hline
\end{tabular}
\caption{Each subset consists of 15GB text.}
\label{sent-doc}
\end{table}

\subsection{Edge Probing tasks}
\label{appendix:ep}

The tasks are \textbf{const}ituent labeling, part-of-speech tagging (POS), named entity labeling (NER), coreference labeling (coref), semantic role labeling (SRL), \textbf{dep}endency labeling \cite{silveira-etal-2014-gold}, semantic protorole labeling (SPR2) \cite{rudinger-etal-2018-neural}, and \textbf{rel}ation classification \cite{hendrickx-etal-2010-semeval}. The first 5 tasks listed are derived from OntoNotes \cite{AB2/MKJJ2R_2013}.

\begin{table*}
\centering
\begin{tabular}{cccccc}
\hline
Domain & Random & Cynical-0.5\% & Cynical-1\% & Cynical-2\% & Cynical-5\% \\ 
\hline
Pile-CC & 27.44\% & 42.06\% & 42.35\% & 43.03\% & 43.30\% \\
OpenWebText2 & 16.95\% & 32.53\% & 32.20\% & 31.79\% & 31.35\% \\
StackExchange & 15.51\% & 3.65\% & 3.56\% & 3.36\% & 3.39\% \\
PubMed Abstracts & 15.40\% & 5.51\% & 5.58\% & 5.17\% & 4.79\% \\
Wikipedia (en) & 8.90\% & 12.03\% & 11.65\% & 11.24\% & 11.09\% \\
USPTO Backgrounds & 5.84\% & 2.00\% & 2.26\% & 2.47\% & 2.55\% \\
PubMed Central & 2.98\% & 0.19\% & 0.24\% & 0.38\% & 0.53\% \\
FreeLaw & 2.66\% & 0.38\% & 0.51\% & 0.81\% & 1.12\% \\
ArXiv & 1.25\% & 0.05\% & 0.06\% & 0.08\% & 0.12\% \\
NIH ExPorter & 0.94\% & 0.39\% & 0.39\% & 0.37\% & 0.36\% \\
HackerNews & 0.82\% & 0.54\% & 0.55\% & 0.60\% & 0.68\% \\
Enron Emails & 0.49\% & 0.51\% & 0.48\% & 0.46\% & 0.43\% \\
OpenSubtitles & 0.33\% & 0.009\% & 0.02\% & 0.03\% & 0.05\% \\
YoutubeSubtitles & 0.17\% & 0.13\% & 0.13\% & 0.14\% & 0.15\% \\
Books3 & 0.15\% & 0.002\% & 0.004\% & 0.009\% & 0.015\% \\
EuroParl & 0.07\% & 0.01\% & 0.01\% & 0.02\% & 0.024\% \\
Gutenberg (PG-19) & 0.04\% & 0.001\% & 0.002\% & 0.005\% & 0.008\% \\
PhilPapers & 0.03\% & 0.002\% & 0.003\% & 0.008\% & 0.013\% \\
BookCorpus2 & 0.01\% & 0.0005\% & 0.001\% & 0.003\% & 0.005\% \\
Ubuntu IRC & 0.01\%  & 0.006\% & 0.004\% & 0.004\% & 0.003\% \\
\hline
\end{tabular}
\caption{Detailed Domain Distribution for the selection under different sizes. }
\label{detailed-data-distribution}
\end{table*}

\end{document}